\documentclass[12pt]{article}
\usepackage{endfloat}
\usepackage{fullpage}

\usepackage{setspace}
\usepackage{graphicx} 

\begin{document}
\title{On the nature of long-range letter correlations in texts}
%%%%%%%% APA 
%\shorttitle{Long-range letter correlations}
%\affiliation{Unaffiliated.\\Email: manin@pobox.com\\Address: 3127 Bryant St., Palo Alto, CA
%  94306, USA\\Phone: 650-575-1506}
%\note{words, figures.}
%%%%%%%%% end APA
\author{D.~Yu.~Manin, manin\@pobox.com}
\maketitle
\abstract{The origin of long-range letter correlations in natural
  texts is studied using random walk analysis and Jensen--Shannon
  divergence. It is concluded that they result from slow variations in
  letter frequency distribution, which are a consequence of slow
  variations in lexical composition within the text. These
  correlations are preserved by random letter shuffling within a
  moving window. As such, they do reflect structural properties of the
  text, but in a very indirect manner.}
\par

\section{Introduction}
Statistical properties of numerical and symbolic sequences derived
from naturally occurring phenomena are of interest in many different
areas. To name just a few examples, human language texts
\cite{EbelingNeiman95,KokolEtAl99,GilletAusloos08,PavlovEbeling01,AmitEtAl93,MontemurroPury02}, music \cite{PavlovEbeling01,Zanette07}, DNA
sequences \cite{GrosseJensenShannon02,PengEtAl92}, and heartbeat
recordings \cite{PengEtAl01} have been subject to such examinations. There
appears to be a common theme in these studies: the sequences in
question are certainly not ``random'' in some sense, they are produced
by ``complex systems'' (we use quotes here to convey an intuitive,
non-terminological status of these statements), and so their
statistical properties should depart from those of random sequences,
thus revealing the regularities. One of the hopes is that we can find
some general characteristics of information-bearing sequences. If this
is possible, not only would we achieve new understanding of the
systems in question, but also it would be possible to apply the
analysis to systems of uncertain status, such as the non-coding
regions of DNA, to determine whether they carry information or not.

In this work we take a closer look at one particular area where
natural-language texts were found to depart from randomness:
long-range letter correlation. Various authors suggested that such
correlations, observed on distances of $10^3$--$10^4$ characters, are
indicative of stylistic and conceptual (semantic) coherence of the
text \cite{AmitEtAl93}, or, more cautiously, ``are of structural
origin'' \cite{EbelingNeiman95}. We will demonstrate that (1) the
long-range letter correlation arises from slow changes in letter
distribution along the text, (2) which in turn result from slow
changes in lexical composition, (3) the primary role being played by
the more frequent words. 

\section{Random walk transformation}

A popular method proposed in \cite{PengEtAl92} for assessing the
degree of randomness in a numerical sequence $\{x_i\}, 0<i<N$, as it is
usually presented, is to consider its members as sequential steps of a
one-dimensional random walk and calculate the mean-square displacement
as a function of time interval:
\begin{equation}
y_{i,k} = \sum_{j=i}^{i+k} x_j 
\end{equation}
\begin{equation}
F(k) = \langle y_{i, k}^2\rangle _i - \left(\langle y_{i,k}\rangle _i\right)^2 
\end{equation}
where the angle brackets $\langle \ldots\rangle _i$ denote the average
over all initial positions $i$ in the sequence, and $k$ is the
interval length assumed to be much shorter than the total sequence
length $k\ll N$.

Or, if we subtract the mean from the data, $\xi_i = x_i - \langle x\rangle $,
then $F(k)$ becomes the mean-square of the partial sums of the resulting sequence, 
\begin{equation}
F(k) = \langle S_{ik}^2\rangle _i,\ \  S_{ik} = \sum_{j=i}^{i+k}\xi_j
\end{equation}
If each $x_i$ results from an independent trial of a random
variable with variance $\sigma^2$, $F(k) = k\sigma^2$ and thus
grows linearly with $k$. If, on the other hand, there are correlations
in the sequence, i.e. some averages $\langle x_ix_{i+k}\rangle _i$ do not
vanish, the growth of $F(k)$ may depart from linearity. Generally
speaking, power-law growth
\begin{equation}
F(k) \sim k^\alpha
\end{equation}
may indicate fractal structure of some sort in the data sequence. The
quantity $\alpha$ is the H\"older (or Hurst) exponent of order~2.

There are many possible ways to convert a natural text to a numerical
sequence in order to calculate $F(k)$. One can use a binary
representation of characters and consider consecutive bit values in it
\cite{KokolEtAl99}, or assign a numerical value to each letter
\cite{EbelingNeiman95}. One can also work
on the level of words and replace each word by its frequency rank
\cite{MontemurroPury02}, or build a binary sequence where 1 (resp., 0)
corresponds to the transition to a longer (resp., shorter) word or to
a more frequent (resp., less frequent) word
\cite{GilletAusloos08}. Regardless of the method used, the cited
authors found departures from the linear growth of the displacement
function $F(k)$. We will loosely follow the method of
\cite{EbelingNeiman95} here to demonstrate the result. We use one of
the texts analyzed in that work, Herman Melville's magnum opus
{\it Moby Dick} having a respectable volume of about $1.2\cdot 10^6$
letters. For comparison, we also utilize Dickens' {\it David
Copperfield} with over $2\cdot 10^6$ letters. Before processing, the
texts were converted to lowercase and non-alphabetic character
sequences were collapsed to single spaces (i.e., in regular
expression terms, {\tt s/[\^{}a-z]+/ /g}). The resulting character
sequence is the subject of all further analysis.

\begin{figure}[tp]
\caption{Displacement function for individual letters in {\em Moby Dick}.}\label{fig:diffusionLetters}
\begin{center}
\input{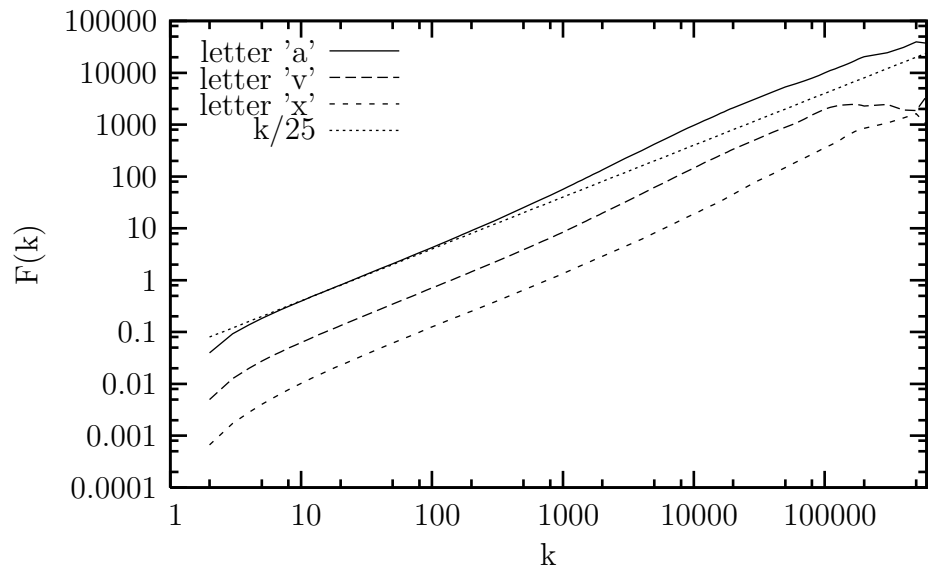}
\end{center}
\end{figure}

To obtain a numerical sequence from the text for the random walk
analysis, following \cite{EbelingNeiman95}, we select a letter and convert all
instances of that letter to ones, and all other characters to zeros.
Fig~\ref{fig:diffusionLetters} shows $F(k)$ for three letters 'a',
'v', and 'x' representative of high, middle, and low frequency
characters.

In the range of $10<k<600000$, where end effects can be neglected,
there are three more of less distinct regions in the chart
Fig.~\ref{fig:diffusionLetters}. Roughly between 10 and 200
characters, $F(k)$ exhibits linear growth indicating lack of
correlations. Between 200 and 1200 characters (for 'a'), it is
consistent with a power-law growth with exponent $\alpha \approx 1.2$
(consistent with the value reported in \cite{EbelingNeiman95}). Above that, $F(k)$
seems to return to linear growth. Note the strikingly similar behavior
of all three letters. It should be noted that the authors of
\cite{EbelingNeiman95} studied the displacement function averaged over
all letters. It is interesting though that long-range correlations are
revealed even in the sequences obtained from individual letters.

This result of \cite{EbelingNeiman95} was corroborated there with
other measures (see also \cite{PavlovEbeling01}, where detrended
fluctuation analysis technique was applied to {\it Moby Dick}). The
question we are concerned with here is what this result actually
means. Ebeling et al. \cite{EbelingNeiman95} demonstrated that if the text is
randomly shuffled --- whether on the level of letters, whole words, or
complete sentences --- the correlations are destroyed, and $F(x)$
returns to the linear growth in the entire valid range of $k$. This
demonstrates that neither intra-word letter correlations, nor
intra-sentence syntactic and semantic relations are responsible for
the observed large-scale behavior. What is, then?

\section{Slow distribution changes}

\begin{figure}[tp]
\caption{Displacement function for a window-shuffled text of {\em Moby Dick}.}\label{fig:windowShuffle}
\begin{center}
\input{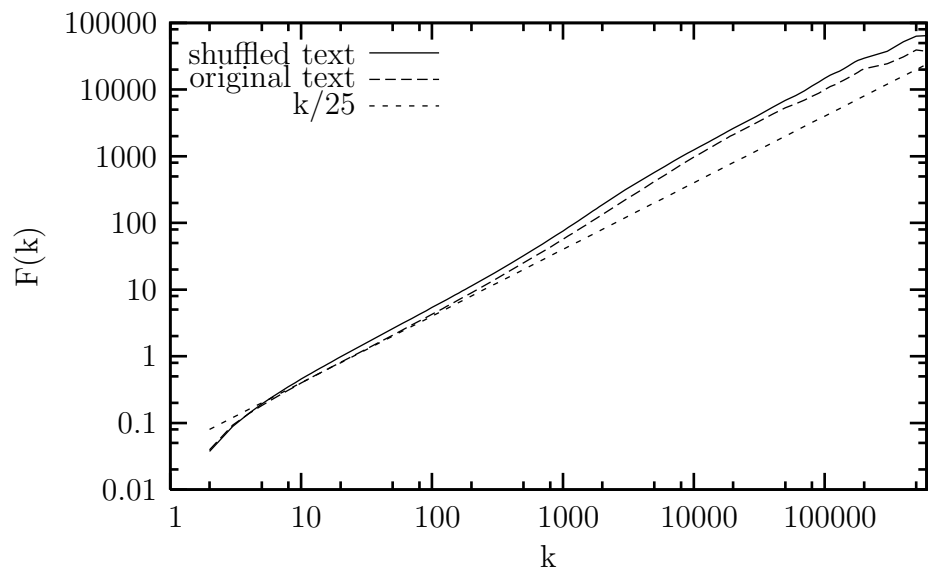}
\end{center}
\end{figure}

The sentence-level shuffling of the text preserves the syntax and
semantics of the language, but destroys the overall narrative with its
plot and composition. Since it also destroys long-range correlation,
it could be tempting to conclude that the correlation is a direct
consequence of the narrative structure. It is easy, however, to
disprove this notion by shuffling the text within a moving
window. Namely, consider the original text as a sequence of characters
$T = \{c_i\}$ and derive from it a character sequence $T'$, where the $i$-th
position is occupied by a character randomly selected from
$\{c_j|i-n/2<j<i+n/2\}$, where $n$ is the window size. This sequence
preserves the overall letter distribution of $T$ and, in addition, any
slow changes in this distribution, but completely destroys everything
else; $T'$ is not a natural language text, and $T$ can not be
reconstructed from it.
Fig.~\ref{fig:windowShuffle} compares the behavior of $F(k)$ for the
original text and for the sequence shuffled with window $n=3000$. Clearly,
all the features of the random walk are preserved by the
shuffling. With increasing window size, as expected, the linear region
extends to the right, and eventually long-range correlations disappear
altogether as window size exceeds the maximum correlation length. 

This leaves little room for speculation: obviously, the specific
behavior of the displacement function is solely a result of bulk
letter distribution in the text, unrelated to any structural features
that distinguish an arbitrary character sequence from a text in a
natural language. 

\begin{figure}[tp]
\caption{Displacement function for an artificial character sequence
  (see text).}\label{fig:artificialSequence}
\begin{center}
\input{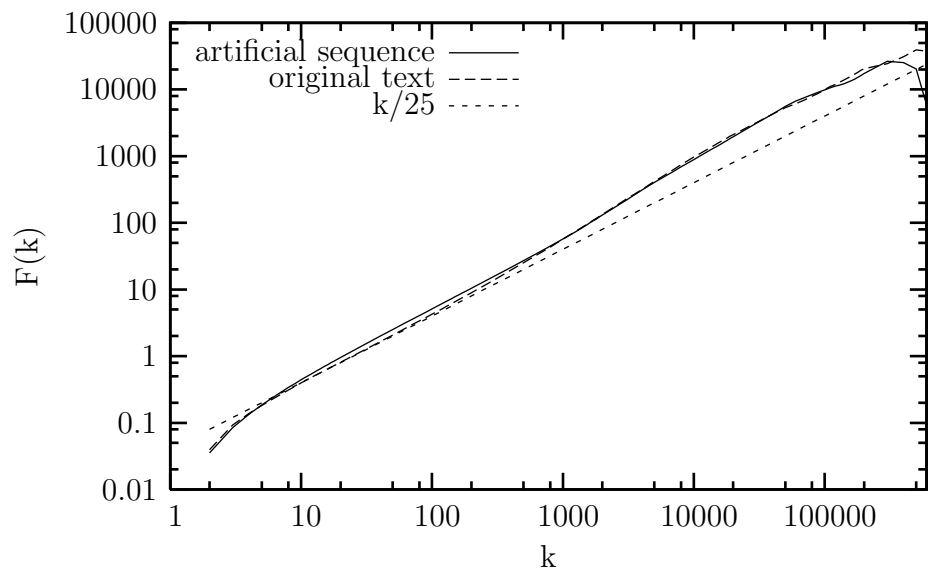}
\end{center}
\end{figure}

To further demonstrate this, we generate another character sequence
which has no relationship to the {\it Moby Dick} text, and results
from a random process that generates the letter 'a' with probability
$p$ and a blank space character with probability $1-p$, where
$p=0.062$ except in a short range of length 6250, where $p=0.1054$
(these parameters were selected to obtain the desired qualitative
behavior of the displacement function; they have no significance
beyond that). Again, the displacement function exhibits the same
qualitative features as for {\it Moby Dick}, as shown in
Fig.~\ref{fig:artificialSequence}. Of course, the distribution of the
letter 'a' in {\it Moby Dick} is very different, but the point is that
$F(k)$ does not reveal the difference.

We can conclude that the behavior of the displacement function $F(k)$
results from slow changes in bulk letter distribution in the text on
the scales of the order $10^3$--$10^4$ characters. But why would the
letter distribution be changing at all? It would seem {\it a priori}
that it should be a rather stable feature of a given language, or at
the very least, a given language subset. For example, children's books
may have a lower frequency of such letters as 'q' or 'x', which in
English appear mostly in the ``long'' words of Latin origin. But there
is no apparent reason why the frequency of such a neutral and common
letter like 'a' should be subject to slow fluctuations.

To investigate this issue, we turn to a different tool. 

\section{Jensen--Shannon information divergence}

We want to compare letter distributions in different segments of the
text. A convenient measure for this is provided by Jensen--Shannon
information divergence (JSD) \cite{GrosseJensenShannon02}. Let $p = \{p+i\}$ and $q = \{q_i\}$, $1<i<n$
be two frequency distributions of the same dimensionality $n$, normalized
so that $\sum p_i = \sum q_i = 1$. Define
\begin{equation}
D(p, q) = H((p+q)/2) - (H(p) + H(q))/2 \label{JSD}
\end{equation}
where $H$ is the entropy of the distribution
\begin{equation}
H(p) = \sum_i p_i\log p_i
\end{equation}
and $(p+q)/2$ is a shorthand for the distribution $r$, such that $r_i
= (p_i + q_i)/2$. (If $p$ and $q$ are determined from different
numbers of trials, they should be weighted with corresponding
coefficients in (\ref{JSD}), but we don't need this generalization
here.) This measure is related to the mutual information between the
two distributions, and vanishes iff they are identical. 

JSD was applied in
\cite{GrosseJensenShannon02} to DNA sequences and in \cite{Zanette07} to texts and
music for the purpose of segmentation, i.e. splitting a sequence into
parts maximizing the difference in composition (whether in terms of
``letters'', ``words'', ``keywords'', etc). Here we have a different
application in mind. We want to find out whether the letter
distribution undergoes statistically significant changes along the
text. To this end, we will compare two adjacent, equal length, regions
of the text, and we need to determine whether the two observed frequency
distributions in them are likely to result from the same underlying
probability distribution. Consequently, we need to calculate the
fluctuation level, i.e. the expected JSD between two realizations of
the same probability distribution. General statistical properties of JSD were
obtained by Grosse et al. \cite{GrosseJensenShannon02}, and we'll
briefly reproduce the derivation for the particular case at hand. 

Let $p$ be the probability distribution and $q$ the observed frequency
distribution obtained from $N\gg n$ trials. The variance of each
$q_i$ is then $\sigma_i^2 = 1/(p_iN)$. Assuming that it is small,
$\sigma_i \ll p_i$, we can represent $q_i = p_i(1+\epsilon_i)$,
$\epsilon_i = O((p_iN)^{-1/2})$ and estimate for each term of the sum
in (\ref{JSD})
\begin{eqnarray}
D_i(p, q) &= \frac{1}{2}\left(p_i\log p_i + q_i\log q_i - (p_i+q_i)\log\frac{p_i+q_i}{2}\right)\\
 &= \frac{p}{2}\left((1+\epsilon_i)\log(1+\epsilon_i) - (2+\epsilon_i)\log(1+\epsilon_i/2)\right)\\
 &\sim \frac{1}{8}p_i\epsilon_i^2 \\
 &=O(1/8N)
\end{eqnarray}
where the first two terms in the Taylor expansion of $\log(1+x)$ were
used (assuming natural logarithms to simplify the expressions). Since
the deviation here is unidirectional, i.e. JSD can not be negative,
the estimate for the sum in (\ref{JSD}) is to be multiplied by $n-1$,
the number of degrees of freedom. Finally, if both $p$ and $q$ are
realizations of an unknown probability distribution, this adds another
factor of 2, and we arrive at
\begin{equation}
{\rm Fluctuation\ level\ of\ } D(p,q) = \frac{n-1}{4N}
\label{JSDfluct}
\end{equation}
where, again, $N$ is the total number of trials in each of $p$, $q$,
and $n$ is the number of possible outcomes\footnote{Interestingly, the
probabilities themselves do not enter this estimate at all. This is in
an apparent contradiction with the fact that a pdf with some number of
possible outcomes $n$ and $p_n=0$ is completely equivalent to a pdf
with $n-1$ possible outcomes, while the estimate (\ref{JSDfluct}) will
be different for them. However the estimate is not valid when some
$p_i$ tends to zero, because this violates the assumption of small
$\sigma_i^2$.}. 

\begin{figure}[tp]
\caption{Jensen---Shannon divergence between letter distributions in
  adjacent segments of length $n$ from {\it Moby Dick}, normalized by
  fluctuation level.}\label{fig:segmentation}
\begin{center}
\input{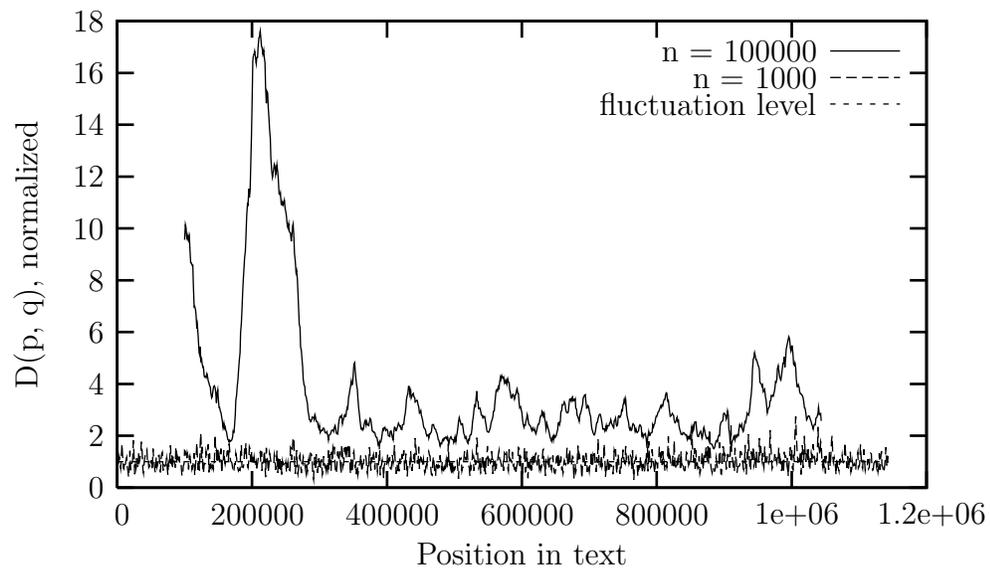}
\end{center}
\end{figure}

Fig.~\ref{fig:segmentation} shows how JSD between adjacent segments of
length $n$ varies along the text for two window sizes, $n = 1000$ and
100000 characters (intermediate sizes not shown to avoid clutter). For
the shortest window size, JSD is at the fluctuation level (which
confirms the estimate (\ref{JSDfluct}) as a side effect). With larger
window size, however, systematic variations in letter composition
stand out from the decreasing statistical noise and become
significant. It may be interesting to see whether the peaks in the
figure match some compositionally meaningful locations in the text,
but for the purposes of this work what's important is that JSD is
comfortably above the fluctuation level practically everywhere, albeit
in some places more so than in others.

Obviously, in the natural text, letters come in packages --- words ---
and any changes in letter composition along the text must result from
the changes in lexical composition. It is well known that words in the
language are distributed in a highly skewed fashion, with many
instances of a small number of frequent word types and increasingly
larger number of rare word types. The distribution is approximately
described by Zipf's law \cite{Zipf49}
\begin{equation}
f_k \sim 1/k
\end{equation}
where $f_k$ is the frequency of the word with rank $k$, and the rank
is the word's sequence number in a dictionary where words are ordered
by decreasing frequency. The top positions in such a dictionary are
occupied by grammatical words (articles, prepositions, personal
pronouns, conjunctions, etc.) and high-frequency significant words
(nouns like {\it man}, adjectives like {\it old}), which are common
for all texts, and by select ``content words'' peculiar to a
particular text ({\it ship, Ahab, whale} in {\it Moby Dick}). The top
of the dictionary is relatively stable, while the rest of it is much
more subject to changes from text to text and within texts, depending
on style, topic, etc. It is not clear {\it a priori} which part(s) of
the lexicon are responsible for the changes in letter composition of
the text: the less frequent words are, generally speaking, more
variable, but because there are many more of them, the law of large
numbers should ensure a more random mixing of the letters; the more
frequent words are less variable, but any change in their distribution
would have a larger impact, because there is a small number of
frequent word types. In the next section we focus on this question.

\section{Lexical composition and its impact on letter distribution}

\begin{figure}[tp]
\caption{Average JSD between two adjacent subsequences of length
  $10^5$ characters for {\it Moby Dick} and {\it David Copperfield}, normalized by
  fluctuation level, calculated with words from 5 frequency ranges.}\label{fig:letterDistrTest}
\begin{center}
\input{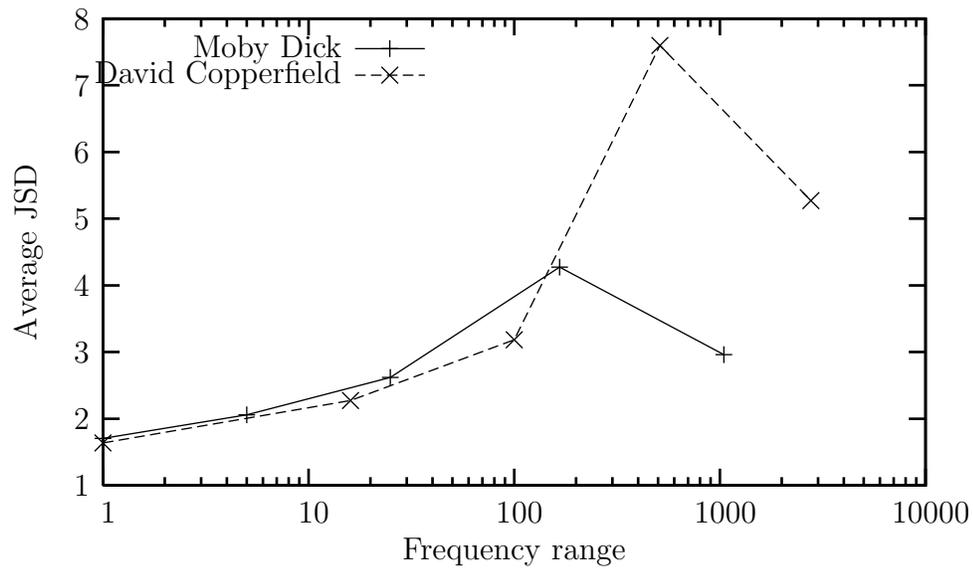}
\end{center}
\end{figure}

To investigate the effect of different parts of the vocabulary, we
applied the analysis of the previous section separately to the words
in different frequency ranges. The frequency dictionary of the text
was subdivided into 5 ranges so that words in each range are
responsible for 20\% of the total number of letters each. For each
range, the rest of the words were blanked out, and JSD between
adjacent 100000-letter segments were calculated (blanks were not
counted). Fig.~\ref{fig:letterDistrTest} shows the average JSD
normalized by the fluctuation level for {\it Moby Dick} and {\it David
Copperfield}. Interestingly, it is somewhat above fluctuation level even for
the most infrequent words, but the biggest contribution in both cases
is due to words that are close to the top of the dictionary, but not the
most frequent ones. It is still a relatively small number of word
types (135 word types for {\it MB} and 80 word types for {\it
DC}). Most names of the major characters fall into this frequency
range. It is a more idiosyncratic set of words than the top of the
lexicon, but it is still limited enough that the letters are not well
mixed according to probabilities.

It is easy to see qualitatively why the slowly changing composition of
the top $\sim 10^2$ words will lead to corresponding changes in letter
distribution and long-range letter correlation. As a simple model,
suppose that 100 ``content'' words are responsible for 20\% of all
letters in a $10^5$-letter text segment. These 100 words are selected
from the lexicon of the language, which, as a whole, is characterized
by some letter frequency distribution $p_i$. The content words are
selected by the writer according to the topic and style, but the
resulting selection of letters is essentially random (except for very
rare cases of highly alliterated prose). However due to the small
number of ``trials'', it will have a considerable variance. For
example, the average frequency of the letter {\it a} in English is
about 10\%, hence out of the $100\cdot 4.5=450$ letters in the 100
content words there will be about 45 'a's. Depending on which 100
words are chosen, the expected variance is on the order of $\sqrt{45}
\approx 7$, i.e. as much as 15\%. Even if the variance in the
remaining 80\% of the text is negligible, he frequency of 'a's
will fluctuate much stronger than for a Poissonian process on the
characteristic lengths where the ``content words'' are stable.

\section{Discussion}
From the analysis we presented in this paper, it follows that
long-range letter correlation in natural texts results from the 
interplay of the following factors:
\begin{enumerate}
\item{a significant portion of the letters in texts is contributed by
  a relatively small class of ``content'' words with high frequency
  and high variability in the text;}
\item{slow variation in the composition of the ``content'' words
  causes corresponding slow variation in the letter distribution; }
\item{this translates to long-range correlation between letters,
  which is invariant with respect to letter shuffling within sliding
  window of length 3000.}
\end{enumerate}
The variation in lexicon may reflect various properties of the
text. For example, here are some of the differences observed between
the first and the second halves of {\it Moby Dick}:
\begin{enumerate}
\item{increased frequency of the word {\it whale} in the second half
  reflects topical differences;}
\item{the word {\it is} is more frequent than {\it was} in the first
  half, but less frequent in the second half, reflecting the
  difference in narrative structure;}
\item{the ratio of articles {\it the} to {\it a} increases from 2.7 in
  the first half to 3.5 in the second half, which may, for example,
  indicate the trend from general statements to concrete narrative.}
\end{enumerate}
The long-range letter correlations can serve as an indirect and
indiscriminate indicator of slow variations of character frequency
distribution. In natural texts, these variations result from the
corresponding slow variations of lexical composition, which in turn
reflect various structural properties of the text. However in the case
of symbolic and numerical sequences of a different origin, such
variations in and of themselves do not necessarily indicate
``complexity'' or information-bearing nature. 

\pagebreak
\bibliographystyle{unsrt}
\bibliography{all}

\end{document}